\documentclass[nohyperref]{article}

\usepackage{microtype}
\usepackage{graphicx}
\usepackage{subfigure}
\usepackage{booktabs} 

\usepackage{subfigure}
\usepackage{graphicx}
\usepackage{amsmath}
\usepackage{amsthm}

\usepackage{amssymb}
\usepackage{tabu,multirow}
\usepackage{subfiles}
\usepackage{multicol}
\usepackage{pbox}
\usepackage{makecell}
\usepackage{colortbl}

\usepackage{hyperref}

\usepackage[accepted]{icml2022}

\usepackage{amsmath}
\usepackage{amssymb}
\usepackage{mathtools}
\usepackage{amsthm}

\usepackage[capitalize,noabbrev]{cleveref}

\theoremstyle{plain}

\theoremstyle{definition}

\theoremstyle{remark}

\usepackage[textsize=tiny]{todonotes}

\usepackage{arydshln}

\icmltitlerunning{Out-of-Distribution Dynamics In RL}
\begin{document}

\twocolumn[
\icmltitle{Out-of-Distribution Dynamics Detection:\\ RL-Relevant Benchmarks and Results}




\begin{icmlauthorlist}
\icmlauthor{Mohamad H.~Danesh}{osu}
\icmlauthor{Alan Fern}{osu}
\end{icmlauthorlist}

\icmlaffiliation{osu}{School of Electrical Engineering and Computer Science, Oregon State University, Corvallis, OR, USA}
\icmlcorrespondingauthor{Mohamad H.~Danesh}{daneshm@oregonstate.edu}

\icmlkeywords{Implicit Quantile Networks, Recurrent Autoregressive Models, Reinforcement Learning, Anomaly Detection, Out-of-distribution dynamics}

\vskip 0.3in
]



\printAffiliationsAndNotice{}  

\begin{abstract}
We study the problem of out-of-distribution dynamics (OODD) detection, which involves detecting when the dynamics of a temporal process change compared to the training-distribution dynamics. This is relevant to applications in control, reinforcement learning (RL), and multi-variate time-series, where changes to test time dynamics can impact the performance of learning controllers/predictors in unknown ways. This problem is particularly important in the context of deep RL, where learned controllers often overfit to the training environment. Currently, however, there is a lack of established OODD benchmarks for the types of environments commonly used in RL research. Our first contribution is to design a set of OODD benchmarks derived from common RL environments with varying types and intensities of OODD. Our second contribution is to design a strong OODD baseline approach based on recurrent implicit quantile network (RIQN), which monitors autoregressive prediction errors for OODD detection. In addition to RIQN, we introduce and test three other simpler baselines. Our final contribution is to evaluate our baseline approaches on the benchmarks to provide results for future comparison. Code are publicly available on GitHub\footnote{The benchmark code is available at:\\ \href{https://github.com/modanesh/anomalous_rl_envs}{https://github.com/modanesh/anomalous\_rl\_envs}\\ The baseline code at:\\ \href{https://github.com/modanesh/recurrent_implicit_quantile_networks}{https://github.com/modanesh/recurrent\_implicit\_quantile\_networks}}.
\end{abstract}

\section{Introduction}\label{sec:intro}
In many machine learning applications, the environment may change in various ways between training and testing. For example, in RL, real-world training can be costly, necessitating simulation and transferring to the real world. Even when RL training and/or validation is possible in the real environment, properties of the environment may change after deployment, e.g., due to mechanical wear, weather conditions, or the behavior of other actors. When such differences between training and deployment arise, a trained policy can behave unpredictably, which is undesirable or even dangerous in many applications. 

One approach for addressing environment change is to follow a robust control framework, for example \citet{abdallah1991survey, sage1999robust}, where the aim is to be robust to certain classes of variations. However, in reality, environments can change in unknown ways that cannot be anticipated at design time. Thus, it is important to be able to detect such changes during operation so that appropriate measures can be taken to ensure safety and avoid large degradation in performance. 

In this paper, we focus on detecting changes in environment dynamics, a problem we refer to as out-of-distribution dynamics (OODD) detection. Specifically, motivated by the challenges of deploying RL controllers, we consider OODD for trained RL controllers operating in common RL-benchmark environments. Our first contribution is to create a set of OODD benchmarks, which contain trajectories under nominal and anomalous dynamics, where different anomalies that effect the dynamics and observations are introduced at random times. The objective is to quickly detect when an anomaly with respect to the nominal trajectories is introduced while maintaining a low false-alarm rate.  

Our second contribution is to develop a baseline set of OODD detection methods and conduct an evaluation on our benchmarks. Our baselines follow the principle of ``anomaly detection by prediction failure", where anomaly signals are generated by comparing nominal predictions of future observations to the actual future observations. Varying the horizon used for future predictions can impact the types of anomalies that can be effectively detected as well as the delay in detection. The baselines differ in the choice of temporal prediction model, which includes simple random forests \cite{breiman2001random}, deterministic recurrent neural networks, as well as a more sophisticated probabilistic model based on recurrent implicit quantile networks (RIQNs).

We evaluate the baselines on our benchmarks using three metrics: AUC, detection delay, and false alarm rate. Overall, some baselines show non-trivial performance and clear advantages over other baselines. However, there appears to be significant room for improving the metrics on the current benchmarks. We hope that this work will inspire new OODD approaches and extensions to the benchmarks and evaluation metrics/protocols. All of our benchmarks and source for baselines and evaluation protocols will be distributed on GitHub after the review period. Currently these materials can be accessed anonymously at  \href{https://drive.google.com/file/d/1APZY-F1J1HVc_vKtEc5yyVNccpAAmHFz/view?usp=sharing}{Benchmarking OODD}.

\section{RL-Relevant OODD Benchmarks}\label{sec:benchmark}

In this section, we describe the construction of our benchmarks, which includes the creation of both nominal (10K) and anomalous trajectories (1K) for each benchmark environment and anomaly type. All of our environments are modeled in the framework of Markov Decision Processes (MDPs) for which we assume basic familiarity.

\subsection{Nominal Trajectory Creation} 

The OODD benchmarks are all derived from common RL benchmark environments. These include three OpenAI Gym classic control environments (Acrobot, Cartpole, and LunarLander), and four Bullet physics environments (Ant, Hopper, HalfCheetah, and Walker2D). 
For all environments, the observations used for OODD are the continuous environment state features, which vary in dimension depending on the environment (see Table \ref{table:anom_groups}), ranging from 4 (Cartpole) to 28 (Ant).
We refer to the each unaltered environment using the default parameters as the \emph{nominal environment}, which is governed by the \emph{nominal dynamics} and \emph{nominal observations}. 

In order to generate environment trajectories that persist for non-trivial durations, we require control policies that achieve non-trivial performance. For this purpose, we train good policies using the IQN algorithm \cite{iqn} for the OpenAI Gym environments and TD3 \cite{fujimoto2018addressing} for the Bullet environments. This set of policies, one per environment, are referred to as the \emph{nominal policies}. 

Finally, for each environment, we generate 10,000 \emph{nominal trajectories}, each a sequence of observations, by executing the the nominal policy in the nominal environment for a finite horizon. These trajectories characterize the behavior of each learned policy under normal environmental conditions, through the lens of observations. If the test time trajectory distribution matches the nominal distribution, then test time performance can be expected to mirror the nominal performance. The goal of OODD is to detect when test-time conditions change by noticing anomalies, which may forecast a possible degregation in policy performance.  


\begin{table*}[t]
\caption{Details of different anomalous environment modifications. (*: for different settings, different agent's components were chosen to be manipulated. From the table, $l$: length, $m$: mass, $mp$: main-engine power, $sp$: side-engine power, and $p$: power.)}
\begin{center}
\begin{footnotesize}
\begin{tabular}{ c|c|c|c|c|c|c|c|c }
\hline\hline
\multirow{2}{*}{Environment} & \multirow{2}{*}{\thead{Observation\\ Dimension}} & \multirow{2}{*}{\thead{IID \\ (mean, std)}} & \multirow{2}{*}{\thead{Calibration\\ Failure}} & \multirow{2}{*}{\thead{Sensor\\ Drift}} & \multicolumn{2}{c|}{Wind} & \multirow{2}{*}{Gravity} & \multirow{2}{*}{Agent's Components*}\\
\cline{6-7}
& & & & & \thead{L2R} & \thead{R2L} & & \\
\hline\hline
\multirow{1}{*}{Acrobot} & 6
& (2, 2) & 10 & 5e-3 & 66\% & 66\% & $[10, 11]$ & $l: [0.5, 2], m: [0.5, 2]$\\
\hline
\multirow{1}{*}{CartPole} & 4
& (1, 2) & 3 & 2e-4 & 66\% & 66\% & $[9, 11]$ & $l: [0, 2], m: [0, 1]$\\
\hline
\multirow{1}{*}{LunarLander} & 8
& (0, 1) & 3 & 1e-4 & 50\% & 50\% & $[9, 10]$ & $mp: [10, 15], sp: [0, 1]$\\
\hline
\multirow{1}{*}{Ant} & 28
& (0, 1) & 2 & 1e-4 & 33\% & 33\% & $[9.8, 10.5]$ & $p: 1.5$\\
\hline
\multirow{1}{*}{HalfCheetah} & 17 
& (0, 0.12) & 1.25 & 2e-5 & 33\% & 33\% & $[9.8, 10.5]$ & $p: 0.6$\\
\hline
\multirow{1}{*}{Hopper} & 15
& (0, 1) & 1.25 & 2e-5 & 33\% & 33\% & $[9.8, 10.5]$ & $p: 0.65$\\
\hline
\multirow{1}{*}{Walker2D} & 22
& (0, 0.12) & 1.25 & 2e-5 & 33\% & 33\% & $[9.8, 10.5]$ & $p: 0.35$\\
\hline\hline
\end{tabular}
\end{footnotesize}
\label{table:noise-details}
\end{center}
\end{table*}

\subsection{Anomalous Trajectory Creation}

For each environment, anomalous trajectories are created by first executing the nominal policy in the nominal environment until a random time point and then switching to an anomalous variant of the environment and continuing executing until a horizon. As described below, we consider two main classes of anomalous variants: \emph{sensor-injected} and \emph{dynamics-injected}. Sensor-injected anomalies correspond to different ways of corrupting the environment observations before sending them to the policy, which can model different types of sensor failure and degradation. Note that such modifications can impact the behavior of the controllers, which in turn can impact the evolution of the trajectories. Dynamics-injected anomalies directly change the dynamics of the environment by modifying  key physical parameters of the environment simulator. 

Our benchmarks include the following 8 types of anomalous variants for each environment--4 sensor-injected and 4 dynamics-injected. Table \ref{table:noise-details} provides relative details for each environment and anomaly type described below.

{\bf IID Noise.} One or more observation sensors is corrupted by IID Gaussian noise. For each environment, the mean and standard deviation are selected as discussed below to avoid near-term failure.

{\bf Sensor Shutdown.} One or more sensors are clamped to zero immediately after the anomaly. 

{\bf Sensor Calibration Failure.} One or more sensors are multiplied by a constant throughout the anomalous run. 

{\bf Sensor Drift.} Relative to the time step in a trajectory, a small amount of noise is injected into the chosen sensor(s). As the trajectory progresses, the magnitude of the injected noise increases. 

{\bf Wind Simulation (L2R and R2L).} Simulated winds starts to blow after injection from either left-to-right (L2R) or right-to-left (R2L), which impacts the environment dynamics. Wind is simulated in the discrete action environments by changing a percentage of the left/right action to noop and reinforcing either the right/left action depending on the direction of the wind. For the continuous-action Bullet environments and appropriate value is added to the action to approximate the impact of wind. 

{\bf Gravity Manipulation.} The value of gravity is changed from the nominal value of 9.8 by a magnitude that depends on the environment.

{\bf Manipulating Components.} Each agent has some physical characteristics, for example, its components' length and masses in Acrobot and CartPole, its engines' powers in LunarLander, or its power in Bullet physics environments. These anomalies randomly change the value of one or more attributes within pre-defined environment-dependent ranges. 

Importantly, we avoid injecting anomalies that cause the nominal policy to completely fail soon after injection (e.g., falling or crashing). This is because such failures are trivial to detect and do not allow testing the ability of detecting anomalies before ``disasters", which is a primary goal. To achieve this, for sensor-injected anomalies, we only modify a small percentage of the features (at most 20\%) chosen for each environment based on tests with the trained controllers. Each anomalous trajectory is generated by randomly selecting the appropriate size subset of features to modify, which provides diversity across the anomalous trajectories. For the dynamics-injected anomalies, we change the physical characteristics of each environment with an environment-relevant value based on pilot tests. Table \ref{table:policy-perf} gives the average reward over 1000 trajectories of the nominal policy in each nominal and anomalous environment. We see a decrease in reward for anomalous environments but not catastrophic levels of decrease. We see that the dynamics-injected anomalies tend to lead to larger performance degradation than the sensor-injected anomalies. 

\section{Prediction-Based OODD Baseline}\label{sec:method}


In this section, we describe a straightforward class of OODD baselines based on the principle of \emph{anomaly detection via prediction error}, which involves the following steps: 1) train a dynamics model that can predict future observations based on past observations, 2) at each time point $t$ use the model to make auto-regressive predictions of observations at time $t+\Delta$, possibly for multiple values of $\Delta$, 3) generate an anomaly signal based on the difference between the predictions and actual observations. Note that when used for real-time anomaly detection, the value of $\Delta$ places a fundamental lower-bound on how much delay there must be before the onset of an anomaly is detected (i.e., a delay of at least $\Delta$). However, for ``slower-moving" anomalies (e.g., subtle sensor drift), it may be necessary to use larger values of $\Delta$ to detect an anomaly's impact. Below we first describe our predictive models, followed by how they generate anomaly signals for OODD.

\begin{table*}[t]
\begin{center}
\caption{Policy performance in different environments with different anomaly types.}
\begin{footnotesize}
\begin{tabular}{ c|c|c|c|c|c|c|c|c|c|c }
\hline\hline
\multirow{2}{*}{Environment} & \multirow{2}{*}{Policy} & \multirow{2}{*}{\thead{Nominal\\ Env}} & \multicolumn{4}{c|}{\thead{Sensor-injected\\ Anomalies}} & \multicolumn{4}{c}{\thead{Dynamics-injected\\ Anomalies}}\\
\cline{4-11}
& & & IID & \thead{Calibration\\ Failure} & \thead{Sensor\\ Drift} & \thead{Sensor\\ Shutdown} & \thead{L2R} & \thead{R2L} & \thead{Gravity} & \thead{Agent's\\ Components}\\
\hline\hline
\multirow{1}{*}{Acrobot}
& IQN & -78.7 & -90.9 & -93.3 & -99.2 & -91.8 & -110.6 & -111.6 & -107.2 & -113.7
\\
\hline
\multirow{1}{*}{CartPole}
& IQN & 500 & 448.3 & 423.8 & 438.0 & 429.1 & 408.7 & 409.3 & 405.4 & 411.7
\\
\hline
\multirow{1}{*}{\shortstack{LunarLander}}
& IQN & 213.5 & 185.8 & 148.3 & 129.3 & 124.4 & 117.1 & 117.8 & 100.5 & 81.0
\\
\hline
\hline
\multirow{1}{*}{Ant}
& TD3 & 3297.2 & 2172.3 & 2190.3 & 2185.5 & 2195.0 & 2022.8 & 2006.9 & 1955.1 & 2051.3
\\
\hline
\multirow{1}{*}{HalfCheetah}
& TD3 & 2820.9 & 2100.6 & 2151.2 & 2002.0 & 2027.6 & 2008.9 & 2026.7 & 2036.7 & 1947.1
\\
\hline
\multirow{1}{*}{Hopper}
& TD3 & 2678.0 & 1928.4 & 1951.3 & 1777.3 & 1928.8 & 1538.3 & 1546.4 & 1474.3 & 1597.2
\\
\hline
\multirow{1}{*}{Walker2D}
& TD3 & 2240.8 & 1665.3 & 1688.5 & 1707.5 & 1732.0 & 1458.4 & 1458.4 & 1513.0 & 1372.1
\\
\hline
\hline
\end{tabular}
\end{footnotesize}
\label{table:policy-perf}
\end{center}
\end{table*}

\subsection{Dynamics Prediction Models}

{\bf Recurrent Implicit Quantile Networks.} To capture aleatoric uncertainty, we ideally want a model that provides a distribution over future observations rather than just a point estimate. For this purpose, we draw on Implicit Quantile Networks (IQNs) \cite{iqn}, which were proposed to represent value-distributions in distributional RL. IQNs learn an implicit representation of a value or feature distribution using the quantile Huber loss. At a functional level, once trained, it can be used to generate samples from the learned distribution conditioned on the input. We refer the reader to the original paper for detailed coverage of IQN. Our model is a recurrent variant of IQN (RIQN) and Figure \ref{fig:model} illustrates the architectural differences compared to IQN with additional details in Appendix \ref{sec:app_riqn}. At a high-level RIQN differs from IQN in three ways: 1) Since we are interested in dynamics prediction in problems where observations may not  fully capture the state (e.g. nominal policies can have internal memory), we add recurrent memory to the IQN network via a GRU \cite{cho2014learning} layer. We found this to be an important extension, which is evaluated in Appendix \ref{sec:app_memory} by a memoryless version of the model. 2) Rather than predicting action values from state-action pairs as input, we train an RIQN for each observation feature to predict the feature distribution at time $t$ based on the previous values of the features. 3) Rather than training via noisy RL-derived target values, we follow a supervised learning approach. 

\begin{figure}[t]
\begin{center}
\subfigure[]{\raisebox{3.5mm}{
\includegraphics[width=\columnwidth]{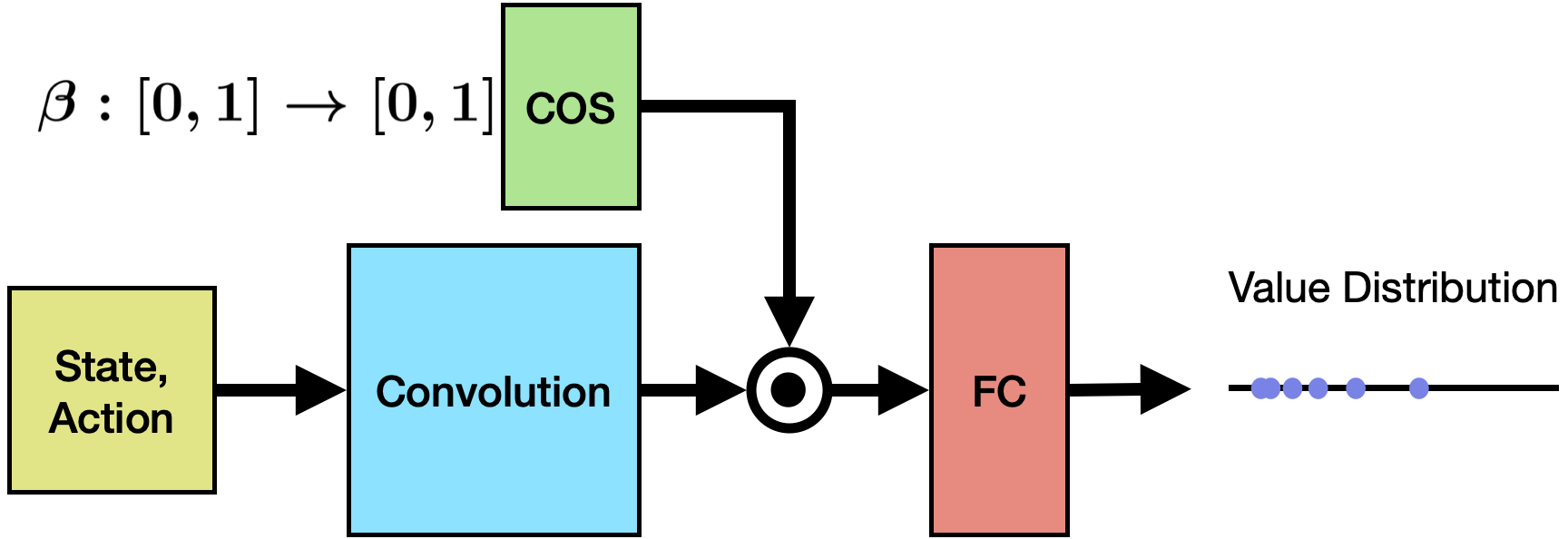}}}
\subfigure[]{
\includegraphics[width=\columnwidth]{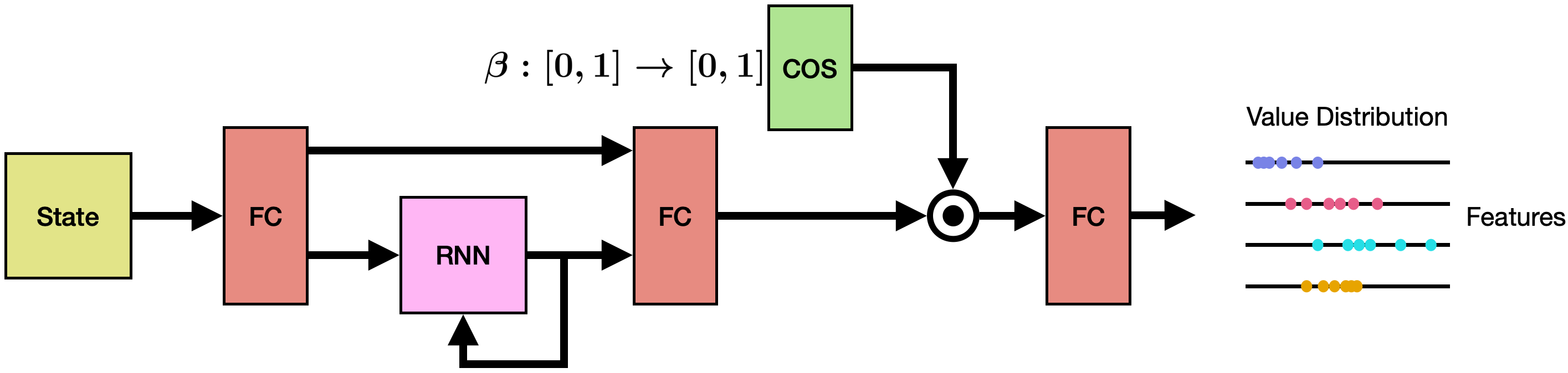}}
\end{center}
\caption{Comparing (a) IQN and (b) RIQ network architectures.}
\label{fig:model}
\end{figure}

More specifically, for supervised learning, we minimize an asymmetric variant of the Huber loss (from prior work \cite{iqn}) on predictions of each feature value at time $t$ given all observation before time $t$. In addition, since we use the RIQN to generate auto-regressive predictions of future observation sequences (see below), we potentially face the problem of compounding prediction error. That is, since the network's outputs are fed back to the input during auto-regressive prediction, error accumulation can occur at test time. To address this, we use the Scheduled Sampling approach \cite{bengio2015scheduled}, which conducts training using inputs from ground truth and auto-regressive samples. 

For anomaly detection, when $\Delta > 1$, we use the RIQN to generate predicted future observations for $\Delta$ time steps. At time step $t$ given prior observations $o_{1:t}$, we consider two ways to generate a future prediction sequence $\hat{o}_{t+1:t+\Delta}$. First, use the RIQN to sample $\hat{o}_{t+1}$ given $o_{1:t}$, followed by sampling $\hat{o}_{t+2}$ given $o_{1:t};\hat{o}_{t+1}$, and so on until sampling a value for $\hat{o}_{t+\Delta}$. By repeating this process for RIQNs, we can generate multiple sampled sequences that for any time $t+\Delta$ yields a set of samples $\{\hat{o}^{(i)}_{t+\Delta}\}$ that can be used to represent the predictive distribution at time $t+\Delta$. 

The second approach we consider is referred to as \emph{mean sampling (RIQN+MS)}, which is intended as a lower variance alternative to pure auto-regressive sampling. This proceeds like regular auto-regressive sampling, but at each step $t+k$ a set of $M$ samples for $t+k+1$ are generated. Next the mean of those samples is computed and used as the next observation for auto-regressive sampling of time step $t+k+2$. This approach also yields a set of observation samples for each time-step, but typically has less variance due to collapsing to the mean during sampling.

{\bf Non-Probabilistic Networks.} In contrast to RIQNs, non-probabilistic networks (NPNs) output a deterministic prediction value at each time step. Since NPNs do not learn distributional representations implicitly, they can be learned using a simple mean-squared loss. Our NPNs have a similar recurrent architecture to the RIQNs, except that they output only a deterministic value given an input. This is achieved by removing the last fully-connected component and the cos transform from the RIQN in Figure \ref{fig:model}b. NPNs can be used to generate deterministic predictions of future trajectories via auto-regressive sampling.  

{\bf Random Forest.} We also use the well-known random forest model \cite{breiman2001random} for predicting the environment dynamics. The model is trained to take a history window of recent observations and predict the next observation. As with NPNs future trajectories can be predicted through auto-regressive sampling.

\begin{figure*}[t]
\begin{center}
\includegraphics[width=\textwidth]{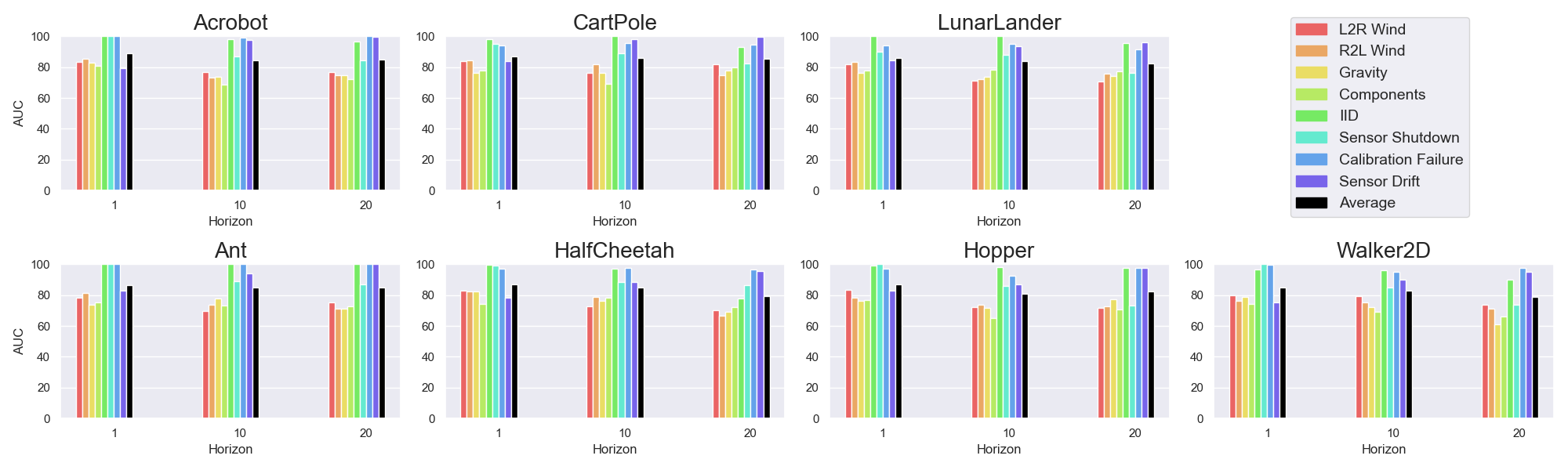}
\end{center}
\caption{How changing the prediction horizon $\Delta$ affects anomaly detection in terms of AUC using the RIQN predictor (without mean sampling).}
\label{fig:compareH}
\end{figure*}

\subsection{Ensembles for Generating Anomaly Signals} \label{sec:ad}

The above RIQN model accounts for aleatoric uncertainty, while none of the above models directly account for epistemic uncertainty. For purposes of anomaly detection, it can be argued that epistemic uncertainty is critical, since it more directly captures out-of-distribution observation sequences. We consider a common approach to addressing this issue by using ensembles of the above models instead of single models. Each model (either RIQN, NPN, or RF) is trained starting from different initial conditions and different hyperparameter settings, attempting to encourage diversity. In this work, we did not attempt to optimize the ensemble training approach, nor did we attempt to quantify how well the ensembles capture epistemic uncertainty. 

Given an ensemble of $e$ dynamics models, an observation sequence $o_{1:H}$, and a detection horizon $\Delta$, we can now define a simple anomaly score to assign to any time $t\in \{\Delta,\ldots, H\}$, denoted by $A_t$. First, use each model model to produce a set of $M$ auto-regressive trajectory samples ($M=1$ for NPN and RF) at time $t$ given the sub-sequence $o_{1:t-\Delta}$, resulting in a set $\{\hat{f}_t^{(1)}, \ldots, \hat{f}_t^{(M\times e)}\}$ of $M \times e$ samples. Given these samples the anomaly score is given by: $A_t = \frac{1}{M\times e}\sum_{i} \left| f_t - \hat{f}_t\right|_1,$ 
which is just the average L1 distance between the samples and the actual observation at time $t$. There are many other ways to define an anomaly score given an ensemble of models, and we leave a deeper comparison for future work. Our initial investigation did not identify an approach significantly better than the above. 

Part of the evaluation in Section \ref{sec:ex} involves evaluating the raw anomaly detection scores (via AUC) and does not require an anomaly detection mechanism. When evaluating detection performance there is a choice of how to combine anomaly scores across time to trigger an alarm. In this work, we use the classic cumulative sum (CUSUM) method \cite{page1954continuous} for this purpose. We used a CUSUM threshold of 0.01 and drift of 0.0018. The threshold sets the amplitude value for the change in the data, and the drift term prevents any change detection in the absence of change.

\section{Experimental Results}\label{sec:ex}



Our experiments aim to address the following questions: 1) What is the overall baseline performance of our RIQN ensemble on the OODD benchmarks compared to NPNs and classical anomaly detectors? 2) How do the detection horizon $\Delta$ and sampling size of the distribution (in RIQN ensemble) affect the performance? 3) What is the delay in detecting anomalies? 4) How much does recurrent memory impact the performance? 5) How do ensembles impact performance? 6) How do different groups of anomalies, sensor-injected and dynamics-injected, affect the performance? 
For space reasons the analysis of sampling size is provided in Appendix \ref{sec:app_ss}. 

Unless otherwise specified, in our experiments, we have an ensemble of 5 models and a sampling size of $M=8$ per RIQN model in the ensemble. As already mentioned, each benchmark data set (one per environment and anomaly type) contains 10K nominal trajectories for training and 1K anomalous test trajectories for evaluating OODD performance. 


\subsection{AUC Performance} 

A common anomaly-detection metric is the \emph{area under the ROC curve (AUC)}, which is independent of detection thresholds and depends only on the quality of the anomaly scores of a detector. Here we evaluate the anomaly scores of our OODD baselines via AUC. To do this, for each test trajectory we calculate the anomaly scores for each time point. Since each an anomaly is injected at a random time in each test sequence and persists to the end, we can also label each time point by whether it is nominal or anomalous. This allows for computing the AUC for each test sequence and then reporting the average AUC across all sequences. 

{\bf Overall AUC.} Figure \ref{fig:compareH} shows results using RIQNs for detection horizons $\Delta=1, 10, 20$ for the different environments and different anomaly types. We see that, with respect to AUC averaged over anomaly types (last bar in each group), the performance for different horizons ($\Delta=1, 10, 20$) are quite similar and that the baseline RIQN approach yields non-trivial overall performance across the benchmarks. However, for the majority of anomaly types it appears that there is significant room for improvement in future work. 

{\bf Impact of Horizon.} Here, we consider the performance RIQN for the individual types of anomalies when the horizon $\Delta$ is changed. First, consider sensor-injected anomalies, where the sensor-drift variant is much more slowly evolving than the other types. This suggests the expectation that longer horizons would be more advantageous for sensor-drift compared to the others. The results in Figure \ref{fig:compareH} agree with this expectation showing that for $\Delta=1$ sensor-drift leads to significantly worse AUC compared to other sensor anomalies. Further, we see that increasing $\Delta$ significantly improves performance for sensor drift. In contrast, the other sensor-injected anomalies all have immediate impacts and in the vast majority of environments the AUC is best at $\Delta=1$. Indeed we see that for the sensor shutdown anomaly, the performance degrades significantly for larger values of $\Delta$. This can be partly explained by the fact that larger values of $\Delta$ cause a smearing effect of the anomaly score, which can decrease the AUC. These results indicate the importance of varying the qualitative types of anomalies in OODD benchmarks to highlight differences in approaches and parameters. 

Regarding the dynamics-injected anomalies, there is less variation in AUC with different horizons $\Delta$. However, there does appear to be a small advantage to using $\Delta=1$ for these anomalies. This indicates that the dynamics anomalies we have introduced may have quite sudden impacts that make longer horizons unnecessary. On the other hand, there is significant room to improve the AUC in all cases, which could suggest that our current baselines are unable to leverage the larger $\Delta$ values to improve performance.  

\begin{figure}[t]
\begin{center}
\includegraphics[width=0.7\columnwidth, height=4cm]{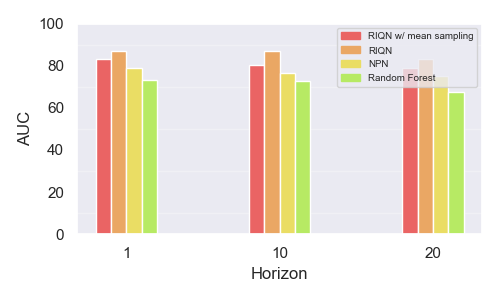}
\end{center}
\vspace{-1em}
\caption{Average AUC of different models across all settings.}
\vspace{-1em}
\label{fig:model_comparison}
\end{figure}

{\bf Comparing Dynamics Models.} Next, Figure \ref{fig:model_comparison} shows the AUC performance of different models (RIQN, RIQN+MS, NN, and Random Forest) averaged over all of the benchmarks. We see that for all horizons $\Delta$ that RIQN (\emph{without} mean sampling) perform the best overall, followed by RIQN+MS, then NPN, and finally Random Forest. The result that RIQN and RIQN+MS outperforms NPN, suggests that there is an advantage to using a probabilistic model that can directly capture aleatoric uncertainty. The observation that RIQN outperforms RIQN+MS indicates that the variance reduction induced by mean sampling has an overall negative impact. The increased diversity of sampled trajectories from RIQN appears to allow for the L1 distances to better discriminate anomalies. The difference between NPN and Random Forest is likely due to the higher expressiveness of the NPN model and also that the NPN model is recurrent and hence can learn to selectively attend to the history. Rather, Random Forest is inherently an atemporal model and requires processing a history window of predetermined size to make its predictions. 
\subsection{Role of memory}\label{sec:app_memory}
The difference between NPN and Random Forests was suggestive that memory is an important element to include in the dynamics models. However, there are too many other confounding difference between those two models. To better observe the impact of memory we trained a model that was identical to RIQN except that: 1) the RNN memory cells were removed, and 2) a history window was provided as input to the model rather than just the current observation. We refer to this model as N-RIQN and a comparison to RIQN is provided in Table \ref{table:memory} for two representative environments averaged across anomaly types. We see that N-RIQN is significantly worse in terms of AUC compared to RIQN for both horizons shown, noting that $\Delta=20$ is qualitatively similar. This provides further evidence that recurrent models with memory are advantageous for OODD. We note that a reasonable effort to improve the performance of N-RIQN by varying certain hyperparmeters was not successful, though future work may identify non-recurrent architectures that exceed the performance achieved in this paper.

\begin{table}[t]
\caption{Comparison of AUCs obtained from RIQN and N-RIQN models. Results are shown for two environments averaged across anomaly types.}
\begin{center}
\begin{footnotesize}
\begin{tabular}{ c|c|c|c|c }
\hline\hline
\multirow{2}{*}{Environment} & \multicolumn{2}{c|}{$\Delta=1$} & \multicolumn{2}{c|}{$\Delta=10$} \\
\cline{2-5}
& \thead{RIQN} & \thead{N-RIQN} & \thead{RIQN} & \thead{N-RIQN}  \\
\hline\hline
\multirow{1}{*}{Acrobot}
& 89.01 & 78.3 & 84.28 & 79.0 \\
\hline
\multirow{1}{*}{Ant}
& 86.5 & 79.8 & 84.7 & 74.7 \\
\hline\hline
\end{tabular}
\end{footnotesize}
\label{table:memory}
\end{center}
\end{table}


{\bf Impact of High-Level Anomaly Types.} In Section \ref{sec:benchmark}, we introduced two primary types of anomalies, sensor-injected and dynamics-injected, which have a total of 8 different anomaly types. From Figure \ref{fig:compareH} we can see a clear trend that for RIQN the dynamics-injected anomalies lead to significantly worse AUCs compared to sensor-injected anomalies. 

For a more quantitative analysis across dynamics-model types, Table \ref{table:anom_groups} gives AUC performance (in addition to other metrics) for the different models when averaged across sensor-injected anomalies and averaged across dynamics injected anomalies. Further, the columns labeled AVG provide the average performance across models for the two high-level types. We see a clear trend that across all environments, for individual models and the averages, the dynamics-injected anomalies lead to significantly worse AUCs compared to sensor-injected anomalies. This suggests that better handling dynamics-injected anomalies and understanding the challenges they pose is a key area of future work in OODD. 

\begin{table}[t]
\caption{AUC results with $\Delta=1, 10$, to study the effect of ensemble size. \emph{\#M} refers to the number of models in the ensemble.}
\begin{center}
\begin{footnotesize}
\begin{tabular}{ c|c|c|c|c }
\hline\hline
\multirow{2}{*}{Environment} & \multicolumn{2}{c|}{$\Delta=1$} & \multicolumn{2}{c}{$\Delta=10$}\\
\cline{2-5}
& \thead{\#M=1} & \thead{\#M=5} & \thead{\#M=1} & \thead{\#M=5} \\
\hline\hline
\multirow{1}{*}{Acrobot}
& 81.5 & 89.01 & 82.2 & 84.28\\
\hline
\multirow{1}{*}{CartPole}
& 83.9 & 86.7 & 83.1 & 85.8\\
\hline
\multirow{1}{*}{\shortstack{LunarLander}}
& 82.2 & 86.0 & 81.1 & 84.1\\
\hline
\hline
\multirow{1}{*}{Ant}
& 83.1 & 86.5 & 79.6 & 84.7\\
\hline
\multirow{1}{*}{HalfCheetah}
& 82.4 & 87.0 & 80.3 & 84.7\\
\hline
\multirow{1}{*}{Hopper}
& 80.2 & 86.7 & 77.6 & 80.8\\
\hline
\multirow{1}{*}{Walker2D}
& 78.1 & 85.1 & 78.5 & 82.8\\
\hline
\hline
\end{tabular}
\end{footnotesize}
\vspace{-1em}
\label{table:ensemble}
\end{center}
\end{table}

{\bf Impact of Ensemble Size.} To study the effect of having an ensemble of models instead of just one model, we experimented an ensemble of five models versus only one model. According to our experiments, we can see that the ensemble perform better regarding the obtained AUC compared to only one model, as presented in Table \ref{table:ensemble}. The reason for such an improvement is because with an ensemble of models, it is easier to address the epistemic uncertainty.

\begin{table*}[t]
\begin{center}
\caption{Comparing AUCs, delays, and FARs with horizon $\Delta=1$ for different anomaly groups. Note: ``RIQN w/ MS" refers to computing the mean between successive steps, and ``RF" is short for Random Forest.}
\begin{footnotesize}
\begin{tabular}{ c|c|c|c|c|c|c|c|c|c|c|c }
\hline\hline
\multirow{2}{*}{Environment} & \multirow{2}{*}{Results} & \multicolumn{5}{c|}{Sensor-injected} & \multicolumn{5}{c}{Dynamics-injected}\\
\cline{3-12}
& & \thead{RIQN} & \thead{RIQN w/ MS} & \thead{NPN} & \thead{RF} & \thead{\bf AVG} & \thead{RIQN} & \thead{RIQN w/ MS} & \thead{NPN} & \thead{RF} & \thead{\bf AVG} \\
\hline\hline
\multirow{3}{*}{Acrobot}
& AUC & 100.0 & 100.0 & 100.0 & 79.16 & {\bf 94.79 } &83.44 & 85.62 & 83.05 & 80.81 & {\bf 83.23 }\\
\cdashline{2-12}
& Delay & 3.38 & 4.47 & 5.79 & 5.86 & {\bf 4.87 } &4.22 & 6.24 & 6.8 & 7.32 & {\bf 6.14 }\\
\cdashline{2-12}
& FAR & 0.09 & 0.1 & 0.1 & 0.11 & {\bf 0.1 } &0.09 & 0.09 & 0.12 & 0.13 & {\bf 0.11 }\\
\hline
\multirow{3}{*}{CartPole}
& AUC & 97.94 & 95.25 & 94.13 & 83.69 & {\bf 92.75 } &83.85 & 84.55 & 76.54 & 78.06 & {\bf 80.75 }\\
\cdashline{2-12}
& Delay & 5.06 & 6.46 & 6.76 & 7.59 & {\bf 6.47 } &3.56 & 7.04 & 7.84 & 8.92 & {\bf 6.84 }\\
\cdashline{2-12}
& FAR & 0.17 & 0.18 & 0.18 & 0.19 & {\bf 0.18 } &0.16 & 0.16 & 0.2 & 0.2 & {\bf 0.18 }\\
\hline
\multirow{3}{*}{\shortstack{LunarLander}}
& AUC & 100.0 & 90.26 & 94.16 & 84.35 & {\bf 92.19 } &81.86 & 83.51 & 76.2 & 77.85 & {\bf 79.85 }\\
\cdashline{2-12}
& Delay & 5.08 & 5.45 & 5.75 & 7.91 & {\bf 6.05 } &6.69 & 7.55 & 7.59 & 8.03 & {\bf 7.47 }\\
\cdashline{2-12}
& FAR & 0.07 & 0.08 & 0.1 & 0.11 & {\bf 0.09 } &0.08 & 0.09 & 0.12 & 0.14 & {\bf 0.1 }\\
\hline
\hline
\multirow{3}{*}{Ant}
& AUC & 100.0 & 100.0 & 100.0 & 83.14 & {\bf 95.78 } &78.31 & 81.56 & 73.88 & 75.16 & {\bf 77.23 }\\
\cdashline{2-12}
& Delay & 4.56 & 4.64 & 5.06 & 5.6 & {\bf 4.96 } &2.58 & 5.18 & 7.09 & 7.96 & {\bf 5.7 }\\
\cdashline{2-12}
& FAR & 0.11 & 0.12 & 0.14 & 0.14 & {\bf 0.13 } &0.07 & 0.12 & 0.14 & 0.15 & {\bf 0.12 }\\
\hline
\multirow{3}{*}{HalfCheetah}
& AUC & 99.53 & 99.3 & 97.01 & 78.29 & {\bf 93.53 } &82.88 & 82.33 & 82.46 & 74.39 & {\bf 80.51 }\\
\cdashline{2-12}
& Delay & 4.34 & 4.61 & 5.58 & 7.14 & {\bf 5.42 } &2.69 & 5.74 & 8.15 & 10.54 & {\bf 6.78 }\\
\cdashline{2-12}
& FAR & 0.12 & 0.12 & 0.12 & 0.13 & {\bf 0.12 } &0.1 & 0.11 & 0.13 & 0.14 & {\bf 0.12 }\\
\hline
\multirow{3}{*}{Hopper}
& AUC & 99.29 & 100.0 & 96.95 & 82.77 & {\bf 94.75 } &83.55 & 78.25 & 76.18 & 76.88 & {\bf 78.72 }\\
\cdashline{2-12}
& Delay & 4.1 & 4.64 & 5.37 & 6.09 & {\bf 5.05 } &5.24 & 7.04 & 7.1 & 7.62 & {\bf 6.75 }\\
\cdashline{2-12}
& FAR & 0.12 & 0.13 & 0.14 & 0.15 & {\bf 0.14 } &0.12 & 0.14 & 0.14 & 0.15 & {\bf 0.14 }\\
\hline
\multirow{3}{*}{Walker2D}
& AUC & 74.03 & 96.82 & 100.0 & 99.59 & {\bf 92.61 } &86.73 & 80.01 & 76.36 & 78.91 & {\bf 80.5 }\\
\cdashline{2-12}
& Delay & 4.52 & 4.61 & 4.67 & 6.22 & {\bf 5.0 } &4.37 & 4.72 & 5.74 & 6.36 & {\bf 5.3 }\\
\cdashline{2-12}
& FAR & 0.12 & 0.13 & 0.14 & 0.15 & {\bf 0.14 } &0.11 & 0.13 & 0.15 & 0.18 & {\bf 0.15 }\\
\hline
\hline
\end{tabular}
\end{footnotesize}
\vspace{-2em}
\label{table:anom_groups}
\end{center}
\end{table*}

\subsection{Anomaly Detection Performance} 

In many applications, it is important not just to provide anomaly scores but also to raise an alarm when an anomalous shift in dynamics is suspected. The alarms should have a minimal \emph{delay}, measured by the number of time steps after injection the alarm is raised, to allow for time-critical interventions when necessary. A small delay, however, must be traded-off with the \emph{false alarm rate (FAR)}, which is the fraction of nominal time points where an alarm is raised. FAR is a traditionally challenging metric in anomaly detection and it is critical have small FARs to support use in practical applications. Here, we use the sequence of anomaly scores for an observation sequence to produce alarms using the classic cumulative sum (CUSUM) method \cite{page1954continuous} (see Section \ref{sec:ad}). 
Quantitative results for the average delay and FAR for each environment averaged across the high-level anomaly types (sensor-injected and dynamics-injected) are reported in Table \ref{table:anom_groups}. In general we see that the FAR rates are not particularly sensitive to the high-level type of anomaly. Given that the AUCs were sensitive, this suggests that false negatives may be dominating the AUC scores. Overall, the FAR values are quite large (on the order of 10\%) for practical applications and improving the FAR metric by orders of magnitude is a critical direction for future work. Further tuning of the CUSUM algorithm may be able to achieve some improvement without significantly impacting other metrics, but it is likely that fundamentally different approaches will be needed. The results for $\Delta>1$ (not shown) are qualitatively similar. 

Comparing delays across anomaly types we see that there is typically slightly more delay for dynamics-injected anomalies, which corresponds to the relative AUC performance. The results for $\Delta>1$ (not shown) are qualitatively similar, except that the delays are typically large by the constant $\Delta$. That is, the inherent increase in delay implied by using a larger $\Delta$ did not payoff in terms of reducing the FAR, AUC, or overall delay.

\vspace{-1em}
\section{Related Work}


Classical anomaly detection algorithms can be categorized into one of the following groups: statistics-based \cite{rousseeuw2005robust}, classification-based \cite{liu2008isolation}, and distance-based \cite{breunig2000lof}. Our work is closest to the distance-based methods. Such methods use the distance of each data point to its nearest neighbor data point or data cluster in order to evaluate how anomalous the given data point is. This group of methods relies on having the appropriate distance criterion for the data. Nevertheless, such algorithms experience problems due to the exponential computational growth in high-dimensional space and can not maintain their efficiency and performance. This is because the computational complexity of the above algorithms is directly related to the increase in data size and the number of samples. Increasing the dimensions leads to scattered cases and makes them less dense \cite{thudumu2020comprehensive}. \citet{aggarwal2017high} suggested that almost any technique based primarily on the concept of proximity is qualitatively degraded in a high-dimensional space and should therefore be defined in a more meaningful way. Although Manhattan and Euclidean distance metrics are among the most used approaches for distance-based anomaly detection, they suffer in high dimensional cases \cite{sadik2014research}. One may address the high dimensionality in data by projecting the dataset into a lower dimension space. It can be done in various ways; for example, principal component analysis or Laplacian eigenmaps are two different approaches to do so \cite{ham2004kernel}.

In addition to classical anomaly detection methods, deep learning-based methods have also gained a lot of attention. These methods can be broadly categorized into: unsupervised, semi-supervised, hybrid \cite{erfani2016high}, and one class learners \cite{chalapathy2018anomaly, scholkopf1999support}. Deep networks are capable of handling high-dimensional data properly. Autoencoders (unsupervised) learn to reconstruct the nominal data; hence, when given an outlier data point, it ends up having a high reconstruction loss because the anomalous data is taken from another distribution most likely \cite{an2015variational}. In this case, the reconstruction loss acts as the anomaly signal. Besides autoencoders, \citet{schlegl2017unsupervised}, \citet{zhai2016deep}, and \citet{zong2018deep} use energy-based models, deep auto-encoding Gaussian mixture models, and GANs to detect anomalies, respectively. In addition to these approaches, recurrent neural network-based methods have flourished in anomaly detection regarding sequential data. One can use RNNs to learn about the nominal time series data in a supervised fashion, then use the learned network as the anomaly detector. \citet{malhotra2015long} proposed to have stacked LSTM-based prediction model. Then, anomalies are detected using the prediction error distribution. \citet{chauhan2015anomaly} used a similar approach to detect various types of injected anomalies existing in the data.

Closest to our approach is \citet{rotman2020online}'s method. They propose a method using which the agent can switch from the learned policy to a safe, heuristic-based policy whenever it realizes that decisions no longer seem reasonable. It is done by measuring uncertainties regarding the observed states, the policy, and the value function. Although their goal is similar to ours, in our method, we use predictive knowledge as a tool to measure the level of uncertainty rather than having an ensemble of policies and storing the history of observed data. We also show the results of our method on more RL-friendly test bases, like OpenAI Gym \cite{brockman2016openai} and Bullet physics  \cite{coumans2021}.

Our approach uses a distance-based method to calculate the anomaly scores while utilizing an RNN-based predictor (RIQN) to take advantage of the underlying correlation in the high-dimensional dataset. The scalability of the RIQN also helps with tackling the curse of dimensionality. Once the autoregressive RIQN has provided the prediction for features at a particular time step, the distance between the true data point and the estimated distribution can be measured. Ideally, for nominal data points, one expects this distance to be small and for anomalous data points to be large, similar to the reconstruction loss in autoencoders. We discuss this further in detail in Section \ref{sec:method}. Overall, to the best of our knowledge, there is no work on detecting anomalies injected into an environment's transition dynamics while the agent is able to perform as well as before. Thus, our proposed method is the first to give RL agents such capability.

\vspace{-1em}
\section{Summary}
In analogy to our work, the area of out-of-distribution detection for image classification has received significant interest in recent years and rapid progress. A key reason for this progress is the availability of OOD benchmarks that allow for easy comparison among methods. A similar set of benchmarks is not yet available for OODD, particularly for environments familiar to RL researchers. Our work has introduced such a set of benchmarks, with the primary goal of setting the stage for follow-up research on OODD approaches, evaluation protocols, and benchmark extensions. In addition, we have proposed several baseline approaches that can be used for comparison and as starting points for future work. While the current baselines achieve non-trivial performance there is significant room for improvement on the current set of benchmarks along multiple metrics. We look forward to the future progress on OODD. 

\bibliography{example_paper}

\begin{thebibliography}{27}
\providecommand{\natexlab}[1]{#1}
\providecommand{\url}[1]{\texttt{#1}}
\expandafter\ifx\csname urlstyle\endcsname\relax
  \providecommand{\doi}[1]{doi: #1}\else
  \providecommand{\doi}{doi: \begingroup \urlstyle{rm}\Url}\fi

\bibitem[Abdallah et~al.(1991)Abdallah, Dawson, Dorato, and
  Jamshidi]{abdallah1991survey}
Abdallah, C., Dawson, D.~M., Dorato, P., and Jamshidi, M.
\newblock Survey of robust control for rigid robots.
\newblock \emph{IEEE Control Systems Magazine}, 11\penalty0 (2):\penalty0
  24--30, 1991.

\bibitem[Aggarwal(2017)]{aggarwal2017high}
Aggarwal, C.~C.
\newblock High-dimensional outlier detection: the subspace method.
\newblock In \emph{Outlier Analysis}, pp.\  149--184. Springer, 2017.

\bibitem[An \& Cho(2015)An and Cho]{an2015variational}
An, J. and Cho, S.
\newblock Variational autoencoder based anomaly detection using reconstruction
  probability.
\newblock \emph{Special Lecture on IE}, 2\penalty0 (1):\penalty0 1--18, 2015.

\bibitem[Bengio et~al.(2015)Bengio, Vinyals, Jaitly, and
  Shazeer]{bengio2015scheduled}
Bengio, S., Vinyals, O., Jaitly, N., and Shazeer, N.
\newblock Scheduled sampling for sequence prediction with recurrent neural
  networks.
\newblock \emph{arXiv preprint arXiv:1506.03099}, 2015.

\bibitem[Breiman(2001)]{breiman2001random}
Breiman, L.
\newblock Random forests.
\newblock \emph{Machine learning}, 45\penalty0 (1):\penalty0 5--32, 2001.

\bibitem[Breunig et~al.(2000)Breunig, Kriegel, Ng, and Sander]{breunig2000lof}
Breunig, M.~M., Kriegel, H.-P., Ng, R.~T., and Sander, J.
\newblock Lof: identifying density-based local outliers.
\newblock In \emph{Proceedings of the 2000 ACM SIGMOD international conference
  on Management of data}, pp.\  93--104, 2000.

\bibitem[Brockman et~al.(2016)Brockman, Cheung, Pettersson, Schneider,
  Schulman, Tang, and Zaremba]{brockman2016openai}
Brockman, G., Cheung, V., Pettersson, L., Schneider, J., Schulman, J., Tang,
  J., and Zaremba, W.
\newblock Openai gym.
\newblock \emph{arXiv preprint arXiv:1606.01540}, 2016.

\bibitem[Chalapathy et~al.(2018)Chalapathy, Menon, and
  Chawla]{chalapathy2018anomaly}
Chalapathy, R., Menon, A.~K., and Chawla, S.
\newblock Anomaly detection using one-class neural networks.
\newblock \emph{arXiv preprint arXiv:1802.06360}, 2018.

\bibitem[Chauhan \& Vig(2015)Chauhan and Vig]{chauhan2015anomaly}
Chauhan, S. and Vig, L.
\newblock Anomaly detection in ecg time signals via deep long short-term memory
  networks.
\newblock In \emph{2015 IEEE International Conference on Data Science and
  Advanced Analytics (DSAA)}, pp.\  1--7. IEEE, 2015.

\bibitem[Cho et~al.(2014)Cho, Van~Merri{\"e}nboer, Gulcehre, Bahdanau,
  Bougares, Schwenk, and Bengio]{cho2014learning}
Cho, K., Van~Merri{\"e}nboer, B., Gulcehre, C., Bahdanau, D., Bougares, F.,
  Schwenk, H., and Bengio, Y.
\newblock Learning phrase representations using rnn encoder-decoder for
  statistical machine translation.
\newblock \emph{arXiv preprint arXiv:1406.1078}, 2014.

\bibitem[Coumans \& Bai(2016--2021)Coumans and Bai]{coumans2021}
Coumans, E. and Bai, Y.
\newblock Pybullet, a python module for physics simulation for games, robotics
  and machine learning.
\newblock \url{http://pybullet.org}, 2016--2021.

\bibitem[Dabney et~al.(2018)Dabney, Ostrovski, Silver, and Munos]{iqn}
Dabney, W., Ostrovski, G., Silver, D., and Munos, R.
\newblock Implicit quantile networks for distributional reinforcement learning.
\newblock \emph{arXiv preprint arXiv:1806.06923}, 2018.

\bibitem[Erfani et~al.(2016)Erfani, Rajasegarar, Karunasekera, and
  Leckie]{erfani2016high}
Erfani, S.~M., Rajasegarar, S., Karunasekera, S., and Leckie, C.
\newblock High-dimensional and large-scale anomaly detection using a linear
  one-class svm with deep learning.
\newblock \emph{Pattern Recognition}, 58:\penalty0 121--134, 2016.

\bibitem[Fujimoto et~al.(2018)Fujimoto, Hoof, and
  Meger]{fujimoto2018addressing}
Fujimoto, S., Hoof, H., and Meger, D.
\newblock Addressing function approximation error in actor-critic methods.
\newblock In \emph{International Conference on Machine Learning}, pp.\
  1587--1596. PMLR, 2018.

\bibitem[Ham et~al.(2004)Ham, Lee, Mika, and Sch{\"o}lkopf]{ham2004kernel}
Ham, J., Lee, D.~D., Mika, S., and Sch{\"o}lkopf, B.
\newblock A kernel view of the dimensionality reduction of manifolds.
\newblock In \emph{Proceedings of the twenty-first international conference on
  Machine learning}, pp.\ ~47, 2004.

\bibitem[Liu et~al.(2008)Liu, Ting, and Zhou]{liu2008isolation}
Liu, F.~T., Ting, K.~M., and Zhou, Z.-H.
\newblock Isolation forest.
\newblock In \emph{2008 Eighth IEEE International Conference on Data Mining},
  pp.\  413--422. IEEE, 2008.

\bibitem[Malhotra et~al.(2015)Malhotra, Vig, Shroff, and
  Agarwal]{malhotra2015long}
Malhotra, P., Vig, L., Shroff, G., and Agarwal, P.
\newblock Long short term memory networks for anomaly detection in time series.
\newblock In \emph{Proceedings}, volume~89, pp.\  89--94. Presses
  universitaires de Louvain, 2015.

\bibitem[Page(1954)]{page1954continuous}
Page, E.~S.
\newblock Continuous inspection schemes.
\newblock \emph{Biometrika}, 41\penalty0 (1/2):\penalty0 100--115, 1954.

\bibitem[Rotman et~al.(2020)Rotman, Schapira, and Tamar]{rotman2020online}
Rotman, N.~H., Schapira, M., and Tamar, A.
\newblock Online safety assurance for deep reinforcement learning.
\newblock \emph{arXiv preprint arXiv:2010.03625}, 2020.

\bibitem[Rousseeuw \& Leroy(2005)Rousseeuw and Leroy]{rousseeuw2005robust}
Rousseeuw, P.~J. and Leroy, A.~M.
\newblock \emph{Robust regression and outlier detection}, volume 589.
\newblock John wiley \& sons, 2005.

\bibitem[Sadik \& Gruenwald(2014)Sadik and Gruenwald]{sadik2014research}
Sadik, S. and Gruenwald, L.
\newblock Research issues in outlier detection for data streams.
\newblock \emph{Acm Sigkdd Explorations Newsletter}, 15\penalty0 (1):\penalty0
  33--40, 2014.

\bibitem[Sage et~al.(1999)Sage, De~Mathelin, and Ostertag]{sage1999robust}
Sage, H., De~Mathelin, M., and Ostertag, E.
\newblock Robust control of robot manipulators: a survey.
\newblock \emph{International Journal of control}, 72\penalty0 (16):\penalty0
  1498--1522, 1999.

\bibitem[Schlegl et~al.(2017)Schlegl, Seeb{\"o}ck, Waldstein, Schmidt-Erfurth,
  and Langs]{schlegl2017unsupervised}
Schlegl, T., Seeb{\"o}ck, P., Waldstein, S.~M., Schmidt-Erfurth, U., and Langs,
  G.
\newblock Unsupervised anomaly detection with generative adversarial networks
  to guide marker discovery.
\newblock In \emph{International conference on information processing in
  medical imaging}, pp.\  146--157. Springer, 2017.

\bibitem[Sch{\"o}lkopf et~al.(1999)Sch{\"o}lkopf, Williamson, Smola,
  Shawe-Taylor, Platt, et~al.]{scholkopf1999support}
Sch{\"o}lkopf, B., Williamson, R.~C., Smola, A.~J., Shawe-Taylor, J., Platt,
  J.~C., et~al.
\newblock Support vector method for novelty detection.
\newblock In \emph{NIPS}, volume~12, pp.\  582--588. Citeseer, 1999.

\bibitem[Thudumu et~al.(2020)Thudumu, Branch, Jin, and
  Singh]{thudumu2020comprehensive}
Thudumu, S., Branch, P., Jin, J., and Singh, J.~J.
\newblock A comprehensive survey of anomaly detection techniques for high
  dimensional big data.
\newblock \emph{Journal of Big Data}, 7\penalty0 (1):\penalty0 1--30, 2020.

\bibitem[Zhai et~al.(2016)Zhai, Cheng, Lu, and Zhang]{zhai2016deep}
Zhai, S., Cheng, Y., Lu, W., and Zhang, Z.
\newblock Deep structured energy based models for anomaly detection.
\newblock In \emph{International Conference on Machine Learning}, pp.\
  1100--1109. PMLR, 2016.

\bibitem[Zong et~al.(2018)Zong, Song, Min, Cheng, Lumezanu, Cho, and
  Chen]{zong2018deep}
Zong, B., Song, Q., Min, M.~R., Cheng, W., Lumezanu, C., Cho, D., and Chen, H.
\newblock Deep autoencoding gaussian mixture model for unsupervised anomaly
  detection.
\newblock In \emph{International Conference on Learning Representations}, 2018.

\end{thebibliography}
\bibliographystyle{icml2022}

\appendix
\onecolumn
\section*{\LARGE{Appendix}}

\begin{figure}[t]
\begin{center}
\includegraphics[width=\columnwidth]{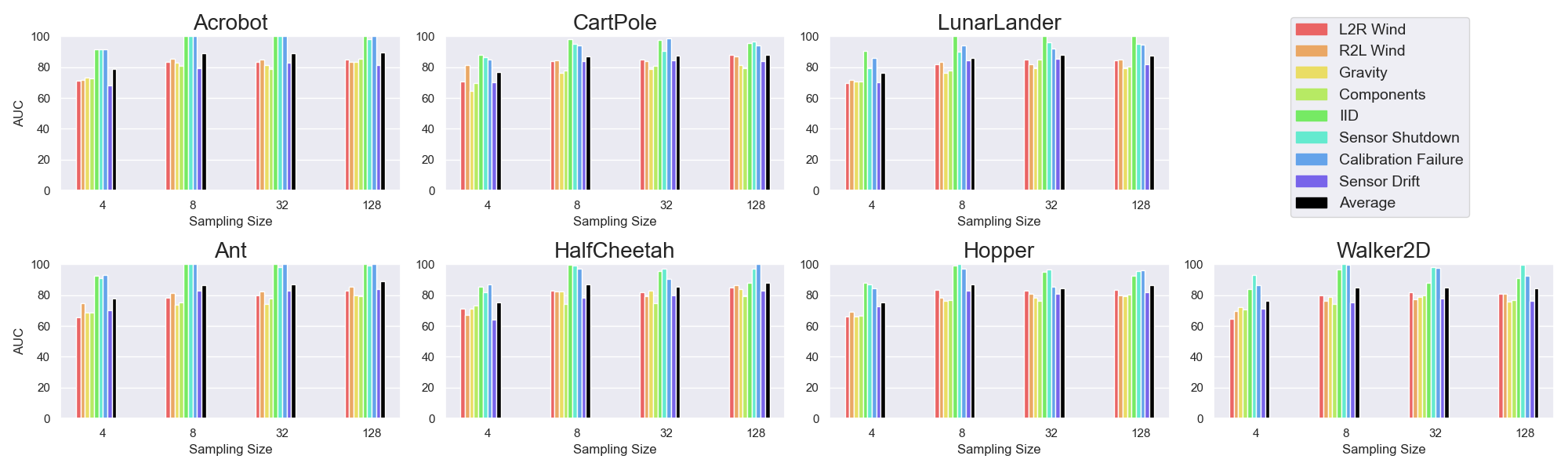}
\end{center}
\caption{The impact of sampling size on anomaly detection for each anomaly type. Horizon is set to 1.}
\label{fig:compareSS}
\end{figure}



\section{Sampling size}\label{sec:app_ss}
The next parameter to investigate is the sampling size, $s$. It basically determines with how many samples we should estimate the true value of a feature in RIQNs. Naturally, one might think the more samples, the better. However, it comes with a larger prediction tree, thus more processing time. Also, a distribution could be represented with a handful of samples. Thus, there would be no need to take a lot of samples. On the other hand, decreasing the number of samples results in a bad empirical estimation and inaccurate predictions. As shown in Figure \ref{fig:compareSS}, having $s=4$ underperforms the case with $s=8, 32, 128$ in all environments. Also, increasing the sampling size to more than 8 gives no absolute benefit while adding processing overhead. This suggests that the true underlying distribution can be almost accurately estimated using 8 samples. Therefore, $s=8$ is considered to be the best choice. It addresses the issues of under-sampling for prediction and processing time.

\section{RIQN Details}\label{sec:app_riqn}
The overall RIQN’s architecture is shown in Figure \ref{fig:model}. It has 64 neurons in the first fully connected layer, 64 memory cells in its memory which is a GRU model \cite{cho2014learning}, and 64 neurons in both the second and the third fully connected layers. Between the first fully connected layer, the GRU, and the second fully connected layer there is a skip connection. The skip connection helps to improve the performance of the network. During the training, we use Adam optimizer with a learning rate of either 0.001 or 0.01.

\end{document}